\documentclass[letterpaper]{article} 
\usepackage{aaai2026}  
\usepackage{times}  
\usepackage{helvet}  
\usepackage{courier}  
\usepackage[hyphens]{url}  
\usepackage{graphicx} 
\usepackage{natbib}  
\usepackage{caption} 
\usepackage{algorithm}
\usepackage{algorithmicx}
\usepackage{algpseudocode}

\usepackage{newfloat}
\usepackage{listings}
\DeclareCaptionStyle{ruled}{labelfont=normalfont,labelsep=colon,strut=off} 
\floatstyle{ruled}
\newfloat{listing}{tb}{lst}{}
\floatname{listing}{Listing}
\title{InstaDA: Augmenting Instance Segmentation Data with Dual-Agent System}
\author{
    Xianbao Hou\textsuperscript{\rm 1,2}\equalcontrib,
    Yonghao He\textsuperscript{\rm 2}\thanks{Project Lead.}\equalcontrib,
    Zeyd Boukhers\textsuperscript{\rm 3},
    John See\textsuperscript{\rm 4},
    Hu Su\textsuperscript{\rm 5},
    Wei Sui\textsuperscript{\rm 2}\thanks{Corresponding authors.},
    Cong Yang\textsuperscript{\rm 1}\footnotemark[3]
}
\affiliations{
    \textsuperscript{\rm 1} Soochow University\\
    \textsuperscript{\rm 2} D-Robotics\\
    \textsuperscript{\rm 3} Fraunhofer FIT\\
    \textsuperscript{\rm 4} Heriot-Watt University (Malaysia)\\
    \textsuperscript{\rm 5} Institute of Automation of Chinese Academy of Sciences\\
    xbhou2024@stu.suda.edu.cn, wei.sui@d-robotics.cc, cong.yang@suda.edu.cn
}

\usepackage{amsmath}
\usepackage{booktabs}
\usepackage{pifont}
\usepackage{tabularx}
\usepackage{makecell}
\usepackage{times}
\usepackage{helvet}
\usepackage{courier}
\usepackage{xcolor}
\newcommand{\cmark}{\ding{51}}%
\newcommand{\xmark}{\ding{55}}%

\usepackage{bibentry}

\begin{document}
\maketitle
\begin{abstract}
Acquiring high-quality instance segmentation data is challenging due to the labor-intensive nature of the annotation process and significant class imbalances within datasets. Recent studies have utilized the integration of Copy-Paste and diffusion models to create more diverse datasets. However, these studies often lack deep collaboration between large language models (LLMs) and diffusion models, and underutilize the rich information within the existing training data. To address these limitations, we propose InstaDA, a novel, training-free \textbf{D}ual-\textbf{A}gent system designed to augment instance segmentation datasets. First, we introduce a Text-Agent (T-Agent) that enhances data diversity through collaboration between LLMs and diffusion models. This agent features a novel Prompt Rethink mechanism, which iteratively refines prompts based on the generated images. This process not only fosters collaboration but also increases image utilization and optimizes the prompts themselves. Additionally, we present an Image-Agent (I-Agent) aimed at enriching the overall data distribution. This agent augments the training set by generating new instances conditioned on the training images. To ensure practicality and efficiency, both agents operate as independent and automated workflows, enhancing usability. Experiments conducted on the LVIS 1.0 validation set indicate that InstaDA achieves significant improvements, with an increase of \textbf{+4.0} in box average precision (AP) and \textbf{+3.3} in mask AP compared to the baseline. Furthermore, it outperforms the leading model, DiverGen, by \textbf{+0.3} in box AP and \textbf{+0.1} in mask AP, with a notable \textbf{+0.7} gain in box AP on common categories and mask AP gains of \textbf{+0.2} on common categories and \textbf{+0.5} on frequent categories. 
\end{abstract}
\section{Introduction}
\begin{figure}[t]
  \centering
  \includegraphics[width=\linewidth]{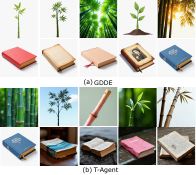}
  \caption{Examples of GDDE proposed by DiverGen~\cite{fan2024divergen} and our T-Agent. Both sets of images are generated by Flux with identical settings. (a) GDDE creates instances with limited visual diversity. (b) The T-Agent generates a significantly more diverse set of instances, showcasing superior variation in object appearance.}
  \label{fig:fig_prompt}
\end{figure}
Instance segmentation~\cite{he2017mask,wang2020solo,cheng2022masked} is a fundamental but challenging computer vision task that requires assigning a unique pixel-level mask to each object instance. The primary bottleneck lies in the prohibitively high labor costs associated with acquiring large-scale, meticulously annotated datasets. Diffusion models have emerged as a promising avenue to mitigate this data scarcity. Recent breakthroughs in text2image models, exemplified by Flux~\cite{flux2024} and SD3.5~\cite{sd3.5}, have significantly advanced the photorealism and fidelity of generated data. Recent studies~\cite{zhao2023x, fan2024divergen} show that data generated by diffusion models can effectively mitigate overfitting and class imbalance when incorporated via the Copy-Paste augmentation~\cite{ghiasi2021simple}. Although these studies thoroughly investigate the application of the Copy-Paste method to generated data and leverage LLMs with diffusion models to enhance data diversity, the potential for deep collaboration between these models remains largely unexplored. Furthermore, high-quality annotated training data is largely an underutilized resource.

To address the aforementioned challenges, we propose a novel dual-agent system comprising two specialized agents: a T-Agent and an I-Agent. The T-Agent leverages deep collaboration between LLMs and diffusion models to generate highly diverse data. Specifically, it employs LLMs in a two-stage process. First, it establishes a comprehensive understanding of prompt diversity by identifying its key visual elements. Second, it leverages these elements to systematically generate novel prompts with the explicit goal of maximizing the diversity of the generated data. To further improve prompt quality, we introduce a Prompt Rethink mechanism that iteratively refines prompts based on feedback from generated images. As illustrated in Fig.~\ref{fig:fig_prompt}, our approach generates significantly more diverse instances by fostering deeper collaboration between models. However, relying solely on the T-Agent for diversity can result in a misaligned data distribution that is suboptimal. To mitigate this, the I-Agent employs ControlNet and image2image~\cite{xlabs2024flux} to augment the training data, ensuring augmented instances align with the original training distribution while enriching data diversity. For practical deployment, both agents are integrated as automated workflows. We further incorporate Low-Rank Adaptation (LoRA)~\cite{hu2022lora,ren2024hypersd,sd35turbox} into the diffusion models to significantly accelerate large-scale image generation.

In each stage of the pipeline, the T-Agent and I-Agent have distinct strategies and objectives. (1) Instance Generation Stage: The T-Agent creates training data based on category labels to maximize diversity. In contrast, the I-Agent generates training data using reference images to enrich the overall data distribution. (2) Foreground Segmentation Stage: Instances from the T-Agent are segmented efficiently using the off-the-shelf model BiRefNet~\cite{zheng2024birefnet}. In terms of instances generated by the I-Agent, we implement our proposed SAM-box strategy. This strategy refines the original masks by using the bounding boxes as prompts for the Segment Anything Model (SAM)~\cite{kirillov2023segment}, ensuring better alignment with the newly created instances. (3) Instance Filtration Stage: We introduce a novel CLIP dual-similarity metric that assesses both image-image (with training data) and image-text similarities to filter out T-Agent instances with significant semantic or visual discrepancies. For the I-Agent, we employ a proportional filtration strategy to retain only the most semantically coherent instances. (4) Instance Augmentation Stage: All high-quality instances generated by both agents are incorporated into the training data via a Copy-Paste method. Outputs from the T-Agent enhance data diversity, while outputs from the I-Agent improve the overall data distribution.

By synergizing the two agents, our method creates a richer and more balanced data distribution, resulting in superior performance on the LVIS~\cite{gupta2019lvis} dataset. In summary, our key contributions are as follows:
\begin{itemize}
\item We present a new dual-agent system consisting of a T-Agent and an I-Agent. The T-Agent utilizes LLMs in a two-stage process and a novel Prompt Rethink mechanism to maximize data diversity. Meanwhile, the I-Agent enhances the overall data distribution by augmenting the training data itself.
\item To ensure the quality of our generated data, we propose a novel CLIP dual-similarity metric that filters instances by enforcing strict, dual thresholds on both semantic and visual similarity, and a SAM-box strategy that leverages bounding boxes as prompts to refine segmentation masks, mitigating annotation misalignments.
\item Experiments show that InstaDA substantially outperforms both the baseline (+4.0 box AP, +3.3 mask AP) and the leading method, DiverGen~\cite{fan2024divergen} (+0.3 box AP, +0.1 mask AP). The gains over DiverGen are particularly notable on common categories (+0.7 box AP, +0.2 mask AP) and frequent categories (+0.5 mask AP).
\end{itemize}

\section{Related Work}
\begin{figure*}[t]
    \centering    \includegraphics[width=\textwidth]{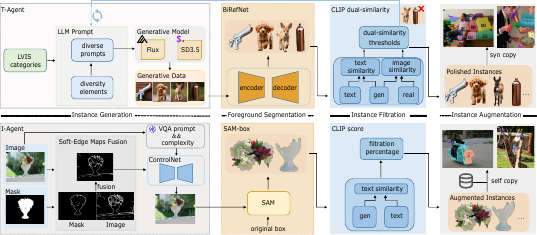} 
    \caption{Overview of the InstaDA pipeline. Our dual-agent system consists of two parallel workflows. The T-Agent (top) first generates images from diverse prompts to enhance data diversity. These instances are then segmented using BiRefNet and filtered by our CLIP dual-similarity metric. Low-quality instances trigger the Prompt Rethink mechanism to refine the initial prompts. These polished instances create a separate synthetic pool for augmentation. In parallel, the I-Agent (bottom) generates instances conditioned on the training images to enrich the overall data distribution. For these instances, our SAM-box strategy is applied for precise annotation, followed by a proportional filtration strategy based on CLIP scores to ensure high quality. These instances form an additional augmentation pool. Finally, instances from both pools are jointly used for Copy-Paste.
    }
    \label{fig_overview}
\end{figure*}
This section reviews previous work that forms the basis for our method, including generative data augmentation and data generation agents.
\subsection{Generative Data Augmentation} 
Generative data augmentation is an effective strategy for improving model performance, especially in situations where labeled data is limited. By utilizing diffusion models, this approach generates new data samples to enhance existing datasets and it has been widely adopted across various tasks, such as image classification~\cite{islam2024diffusemix}, object detection~\cite{fang2024data}, and segmentation~\cite{yang2023freemask,nguyen2023dataset,fan2024divergen,zhao2023x}. 

In segmentation, FreeMask~\cite{yang2023freemask} uses diffusion models to produce high-quality images with dense annotations derived from semantic masks in real datasets. Similarly, Dataset Diffusion~\cite{nguyen2023dataset} employs Stable Diffusion~\cite{rombach2022high} to generate synthetic images along with their corresponding labels based on text prompts. Our work is closely related to methods that integrate diffusion models with the Copy-Paste technique~\cite{ghiasi2021simple}, such as X-Paste~\cite{zhao2023x} and DiverGen~\cite{fan2024divergen}. X-Paste initially demonstrates the effectiveness of overlaying generated instances onto training images; however, it falls short in addressing the diversity of the generated data. DiverGen, in turn, builds on this concept by proposing the GDDE strategy to enhance such diversity. Despite its use of LLMs to generate prompts for diffusion models, its methodology is constrained by simplistic instructions to LLMs and a limited set of diversity elements.

Our approach advances these methods in two key areas: deepening collaboration between LLMs and diffusion models to further improve data diversity, and using training images as conditional references to guide image generation, thereby refining and enriching the data distribution.

\subsection{Data Generation Agent}
A data generation agent is a system designed to automate and manage multiple models, maximizing the capabilities of diffusion models while simplifying complex generation processes. Significant research has been conducted in this area~\cite{chen2023llava, wang2024genartist, li2025mccd}. GenArtist~\cite{wang2024genartist} integrates image generation and editing through a multimodal large language model agent. Similarly, MMCD~\cite{li2025mccd} uses multiple intelligent agents to interpret prompt information and generate images with high prompt fidelity through a hierarchical diffusion process. However, the complex planning and reasoning required by general-purpose agents can make them too time-consuming for large-scale data generation.

To address this issue, we propose a dual-agent system consisting of two specialized agents. This specialization enables our dual-agent system to achieve the efficiency required for large-scale data generation.

\section{Methodology}
This section outlines the methodology of InstaDA, a dual-agent system created to enhance the quality and utility of instance segmentation data, as depicted in Fig.~\ref{fig_overview}. The pipeline consists of four sequential stages: Instance Generation, Foreground Segmentation, Instance Filtration, and Instance Augmentation. We will detail the T-Agent and the I-Agent in sequence, highlighting their unique designs and providing a comprehensive overview of the full pipeline implementation for each.
\subsection{T-Agent}
\paragraph{Instance Generation}
To systematically generate diverse and complex prompts, we introduce a novel two-stage LLM collaboration. Instead of relying on human intuition or manually designed diversity elements, we utilize DeepSeek-R1~\cite{guo2025deepseek} to automatically extract a comprehensive set of elements that influence visual diversity. Based on this rich set of identified elements, we leverage DeepSeek-V3~\cite{liu2024deepseek} to generate complex prompts that combine various elements to target unique visual characteristics. This automated and structured approach addresses the limitations of conventional methods, which often depend on simplistic directives like ``output different prompts" and may overlook significant visual variations.

To maximize the potential of these diverse prompts and ensure better adherence to them, we employ advanced diffusion models, such as Flux~\cite{flux2024} and SD3.5~\cite{sd3.5}, for image generation. Traditional generative pipelines typically discard low-quality images, which leads to wasted potential of feedback. Therefore, we introduce a mechanism called Prompt Rethink. 

This mechanism treats a low-quality generated image as an indicator of a flawed prompt, creating a feedback loop where DeepSeek-V3 iteratively refines the prompt. It employs a two-step fallback process: first, it attempts to correct semantic errors by simplifying the prompt to a standard template, such as "a photo of a single \textit{\{category\_name\}}," while retaining the original diversity elements. If this adjustment fails, DeepSeek-V3 then assesses the diversity elements themselves as unsuitable and generates an entirely new prompt with a more plausible combination. This tiered approach maximizes data utilization and enables the LLM to adapt based on visual feedback from the diffusion models.

\paragraph{Foreground Segmentation}
Achieving an optimal balance between annotation quality and speed is essential for large-scale foreground segmentation. We tackle this by utilizing a single high-performance model, BiRefNet~\cite{zheng2024birefnet}, which ensures annotation quality while avoiding the inefficiencies of running multiple models as seen in X-Paste~\cite{zhao2023x}. We found the significant computational overhead of this multi-model strategy is often unwarranted, as a single top-performing model typically provides the most substantial performance gains. To further improve speed, we optimize BiRefNet with TensorRT~\cite{TensorRT} via an open-source implementation~\cite{bireftrt}, reducing the average inference time from 4.43 seconds to 0.11 seconds.

\paragraph{Instance Filtration}
To effectively filter generated instances, we introduce the CLIP dual-similarity metric, which eliminates instances with significant semantic or visual discrepancies. Our approach calculates two similarity scores for each generated instance: text similarity with a predefined template, and image similarity between the embeddings of training instances in the same category. Both scores are computed using the CLIP encoders~\cite{radford2021learning}. An instance is retained only if it exceeds predefined thresholds for both scores. If an instance fails to meet both thresholds, it is marked as low-quality and triggers the Prompt Rethink mechanism. This metric thus strikes a balance between diversity and domain fidelity while also providing a failure signal that drives the collaborative Prompt Rethink loop.

\paragraph{Instance Augmentation}
We designate instances generated by the T-Agent as the first of two independent synthetic pools. We then apply the Copy-Paste~\cite{ghiasi2021simple} method by sampling instances from this synthetic pool.

\subsection{I-Agent}
\begin{figure}[t]
  \centering
  \includegraphics[width=\linewidth]{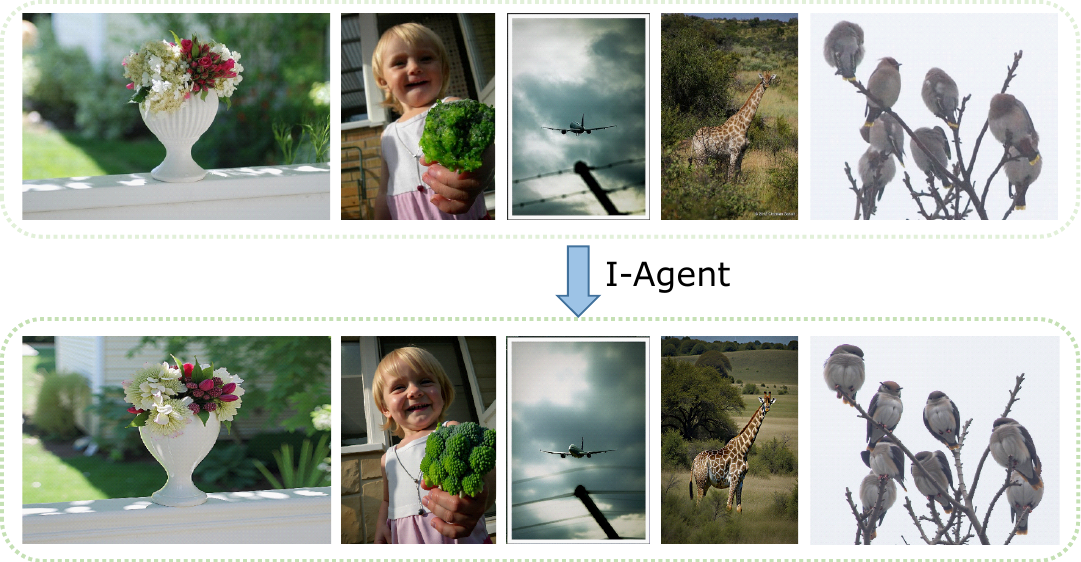}
  \caption{Examples of our proposed I-Agent. Conditioning on a source image (top), our I-Agent generates augmented data (bottom) with subtle yet meaningful alterations to object details. Close inspection reveals these nuanced changes, which serve to enrich the overall data distribution.}
  \label{fig:fig_IAgent}
\end{figure}
\paragraph{Instance Generation}
To exploit the underutilized training data, we introduce the I-Agent to enhance the data distribution by generating images guided by existing training samples. However, for complex, multi-instance images, relying solely on the standard structural guidance of ControlNet~\cite{zhang2023adding} can be insufficient. To address this, we enhance the standard pipeline by incorporating three key modifications: we guide the diffusion model with (1) a fused edge map derived from both the training image and its mask for more robust structural control; (2) a text prompt from Qwen2-VL-7B~\cite{wang2024qwen2} for semantic fidelity; and (3) an initial latent from the training image, encoded by a Variational Autoencoder (VAE)~\cite{kingma2013auto} to replace the random noise. Crucially, the denoising strength is adaptive based on an image complexity score~\cite{feng2023ic9600}. This strategy allows for subtle modifications in complex scenes to preserve their integrity while enabling more significant changes in simpler images to maximize diversity, as shown in Fig.~\ref{fig:fig_IAgent}.

The fused edge map, $E_{\text{fused}}$, is created using Holistically-Nested Edge Detection (HED)~\cite{xie2015holistically} and a weighted combination:
\begin{equation}
E_{\text{fused}} = \alpha \cdot E_{\text{image}} + (1 - \alpha) \cdot E_{\text{mask}}\quad .
\end{equation}
where $E_{\text{image}}$ and $E_{\text{mask}}$ are the soft-edge maps of the image and its corresponding mask, respectively. The parameter $\alpha \in [0, 1]$, empirically set to 0.5, controls the fusion strength.

Algorithm~\ref{alg:i_agent_generation} details our image generation process. It conditions a diffusion model on three guides from a training image and its mask: a latent code for initialization, a fused edge map for structure, and a text prompt for semantics. The entire process is regulated by an adaptive denoising strength.
\begin{algorithm}[t]
\caption{Image Generation for the I-Agent}
\label{alg:i_agent_generation}
\begin{algorithmic}[1]
\State \textbf{Input:} A training image \(I_{train}\) and its corresponding instance mask \(M_{mask}\).
\State \textbf{Output:} An augmented image \(I_{aug}\).
\State \(P_{text} \gets \text{Qwen2-VL-7B}(I_{train})\)
\State \(E_{image} \gets \text{HED}(I_{train})\)
\State \(E_{mask} \gets \text{HED}(M_{mask})\)
\State \(E_{fused} \gets \alpha \cdot E_{image} + (1 - \alpha) \cdot E_{mask}\)
\State \(C \gets \text{ComplexityScore}(I_{train})\)
\State \(Z_{0} \gets \text{VAE.encode}(I_{train})\)
\State \(S_{denoise} \gets \text{AdaptiveStrength}(C)\)
\State \(Z_{final} \gets \mathcal{M}_{diffusion}(\text{initial\_latent}=Z_{0}, \text{prompt}=P_{text}, \text{control}=E_{fused}, \text{strength}=S_{denoise})\)
\State \(I_{aug} \gets \text{VAE.decode}(Z_{final})\)
\State \Return \(I_{aug}\)
\end{algorithmic}
\end{algorithm}

\paragraph{Foreground Segmentation}
To ensure high-quality annotations for the augmented instances, we introduce SAM-box, a novel re-annotation strategy. This strategy is essential because, although our I-Agent generates visually plausible instances, their boundaries may slightly diverge from the original segmentation masks during the generation process. This misalignment in annotations is known to degrade model performance~\cite{fang2024data}. Our SAM-box strategy proactively addresses this issue by leveraging the inherent robustness of bounding boxes. Rather than relying on the potentially misaligned original masks, it uses the bounding box from the original annotation as a spatial prompt for the Segment Anything Model (SAM)~\cite{kirillov2023segment}. Since bounding boxes are less sensitive to minor changes in object shape and pose, they provide a more reliable guide for SAM to generate a new and accurate segmentation mask that adheres closely to the contours of the instance in the newly generated image. This process corrects boundary inaccuracies and significantly enhances the quality of the annotation.
\begin{table}[t]
  \small
  \centering
  \setlength{\tabcolsep}{1.88mm}
      \begin{tabular}{c|c|cccc}
      \toprule
      \
      Gen Data&Method&$\text{AP}^{box}$&$\text{AP}^{mask}$&$\text{AP}_r^{box}$&$\text{AP}_r^{mask}$ \\
      \midrule
      600k & GDDE &  50.67  &  44.99 & \textbf{48.52} & \textbf{43.63} \\
      480k & T-Agent &   \textbf{50.79}  &  \textbf{45.03}&  48.47  & 43.26 \\
    \bottomrule
    \end{tabular}
\vspace{-2mm}
\caption{Results of diversity under different data quantities. Our T-Agent achieves comparable overall performance to the GDDE proposed by DiverGen using only a limited amount of data, which demonstrates the greater diversity of the data generated by our approach.}
\label{tab:compare_divergen}
\vspace{-3mm}
\end{table}
\paragraph{Instance Filtration}
After refining masks with our SAM-box strategy, a subsequent filtration step is needed to address issues beyond annotation quality. Some generated instances, even with accurate masks, may still exhibit low semantic quality. To filter these instances, we employ a proportional filtration strategy inspired by ControlAug~\cite{fang2024data}, rather than relying on a fixed threshold for each instance. This image-level approach is particularly suitable for the I-Agent, which generates complex
scenes with multiple instances. Our objective is to retain a certain percentage of the highest-quality images, as this is more meaningful than maintaining a variable number of individual instances. Our three-step filtration strategy first scores each individual mask $m$. By calculating its percentile score $p_m$, which ranks its CLIP score $s_m$ against others from the same object category $c_m$ within the set $\mathcal{S}_{c_m}$ that consists of all CLIP scores corresponding to category $c_m$.
\begin{equation}
p_m = \frac{|{ s \in \mathcal{S}_{c_m} \mid s \leq s_m }|}{|\mathcal{S}_{c_m}|} \times 100\quad .
\label{eq:mask_percentile}
\end{equation}

Next, the overall quality score for an image $j$, $\bar{p}_j$, is the average percentile score of its constituent masks $\mathcal{M}_j$.
\begin{equation}
\bar{p}_j = \frac{1}{|\mathcal{M}_j|} \sum_{m \in \mathcal{M}_j} p_m\quad .
\label{eq:image_score}
\end{equation}

Finally, to retain the top $k$ percent of the batch, where $k$ is set to 20 in our experiments, an image $j$ is kept only if its score $\bar{p}_j$ is at least the dynamic threshold. This threshold is the $(100-k)$-th percentile of $\mathcal{A}$, where $\mathcal{A}$ is the set containing the overall quality scores ($\bar{p}_j$) of every generated image with at least one mask.
\begin{equation}
\bar{p}_j \ge \text{Percentile}\left( \mathcal{A}, 100 - k \right)\quad .
\label{eq:proportional_filtration}
\end{equation}

\paragraph{Instance Augmentation}
Instances generated by the I-Agent complement those created by the T-Agent. Because they are directly derived from and conditioned on the original LVIS training data, these new instances are uniquely positioned to enhance the source data. Specifically, they are integrated with the original LVIS instances to form an augmented source pool. This pool is used in conjunction with the synthetic pool generated by the T-Agent. During the final Copy-Paste augmentation, instances are selected from both pools to maximize data quality and diversity. This integration strategy enriches the training data while ensuring that the overall distribution remains true to the source domain.

\section{Experiments}
\begin{table}[t]
  \small
  \centering
  \setlength{\tabcolsep}{2mm} 
      \begin{tabular}{c|c|cccc}
      \toprule
      Gen Data&T-Agent&$\text{AP}^{box}$&$\text{AP}^{mask}$&$\text{AP}_r^{box}$&$\text{AP}_r^{mask}$ \\
      \midrule
      0 &  &  47.50  & 42.32  & 41.39  & 36.83 \\
      240k &  &  50.38  &  44.46 & 46.36 & 40.79 \\
      960k &  &  50.26  & 44.50  & 45.89  & 41.05 \\
      1200k &  &  49.85  & 44.24 & 43.52  & 39.14 \\
      240k & \cmark &   50.75  &  44.91 &  48.14  & 42.30 \\
      960k & \cmark & \textbf{51.17}  & 45.16 & \textbf{50.19}  & \textbf{44.52} \\
      1200k & \cmark & 51.01  & \textbf{45.37} & 48.24  & 43.53 \\
    \bottomrule
    \end{tabular}
\vspace{-2mm}
\caption{Analysis of data quantity and diversity. Our T-Agent shows significant performance improvements across all data volumes, and these benefits are especially noticeable at larger scales. The results indicate that having a high level of data diversity is essential for reducing overfitting.}
\label{tab:TAgent}
\vspace{-3mm}
\end{table}
In this section, we conduct a comprehensive set of experiments to validate the effectiveness of our proposed InstaDA. Details on specific hyperparameters, the efficiency analysis of our data generation, and the rationale for selecting proportional filtration are provided in the appendix.
\subsection{Experimental Setup}
\paragraph{Datasets}
The experiments are conducted on the LVIS dataset~\cite{gupta2019lvis}, which uses images from COCO~\cite{lin2014microsoft} but with a revised set of categories and annotations. The dataset is partitioned into official training and validation splits, containing approximately 100k and 20k images, respectively. Its 1,203 categories are divided into rare, common, and frequent groups, and are marked by a significant class imbalance. On average, common categories contain 12 times more instances, and frequent categories 228 times more, than rare categories. We follow the official splits for all training and evaluation.

\paragraph{Evaluation Metrics}
The evaluation metrics are standard LVIS metrics, namely box average precision ($\text{AP}^{box}$) and mask average precision ($\text{AP}^{mask}$). For a more fine-grained analysis, we also report these metrics across the three official LVIS frequency groups of rare, common, and frequent, which are denoted by the subscripts ${\text{r}}$, ${\text{c}}$, and ${\text{f}}$, respectively.

\paragraph{Implementation Details}
Our baseline is CenterNet2~\cite{zhou2021probabilistic} with a Swin-L~\cite{liu2021swin} backbone, implemented in Detectron2~\cite{wu2019detectron2}. Models are trained for 180k iterations with an input image size of 896 and a batch size of 16. We use an exponential moving average (EMA) decay of 0.999 and a base learning rate of 0.0001. For data generation, we employ Flux~\cite{flux2024} and SD3.5~\cite{sd3.5} due to their superior prompt adherence and quality. Both agents are implemented as automated JSON workflows in ComfyUI~\cite{comfyui} for efficiency.
\subsection{Main Results}
\paragraph{Data Diversity Matters More Than Quantity}
By systematically engineering diverse prompts, the T-Agent generates data with demonstrably high diversity. To highlight this, we compare our T-Agent against the GDDE method from DiverGen~\cite{fan2024divergen} in Table~\ref{tab:compare_divergen}. Using only Flux to generate 480K samples, our method surpasses the GDDE method with its 600K data, achieving gains of +0.12 in box AP and +0.04 in mask AP. This indicates our generated data, while smaller in volume, possesses greater diversity, making it more effective at mitigating overfitting. Furthermore, this underscores the critical role of diversity, demonstrating that significant gains in data diversity can yield superior performance even with less generated data.

\paragraph{As Data Quantity Grows, Data Diversity Becomes Increasingly Critical}
\begin{figure}[t]
  \centering
  \includegraphics[width=\linewidth]{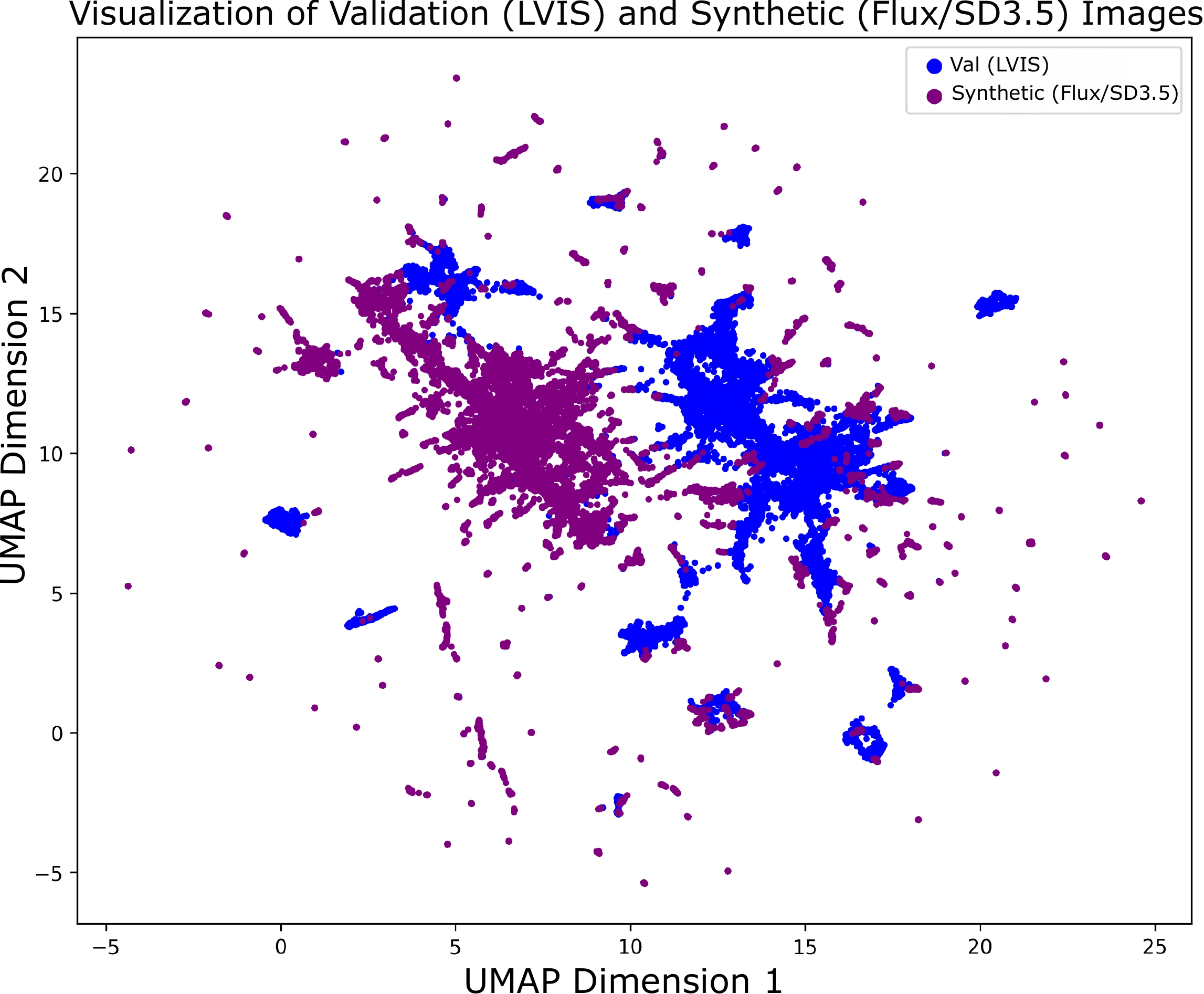}
  \caption{Visualization of data distribution on generated data and LVIS dataset. This UMAP visualization illustrates that excessive volume of generated data creates a suboptimal data distribution relative to the LVIS validation set, leading to degraded model performance.}
  \label{fig:data_distribution}
\end{figure}
We examine the impact of data quantity with experiments on three dataset sizes: 240k, 960k, and 1,200k samples, corresponding to 0.2k, 0.8k, and 1k samples per category, respectively. The data is generated by Flux and SD3.5, with each model contributing half the total volume. As shown in Table~\ref{tab:TAgent}, without the T-Agent, increasing the data from 240k to 960k results in no performance improvement and degrades box average precision (AP). Expanding to 1,200k samples leads to a further decline in performance. This trend indicates significant overfitting. 

In contrast, the T-Agent significantly enhances performance. At the 240k scale, it improves box AP by +0.37 and mask AP by +0.45. This positive impact is more pronounced at the 960k scale, where gains increase to +0.91 and +0.66, respectively. The ability of the T-Agent to systematically enhance data diversity is crucial, and its benefits become more pronounced as data volume increases. This highlights the effectiveness of our approach, emphasizing that highly diverse generated data is essential to mitigate overfitting.

However, increasing the data volume from 960k to 1,200k results in a decline in performance even with the T-Agent. Our UMAP~\cite{mcinnes2018umap}  analysis, shown in Fig.~\ref{fig:data_distribution}, indicates that this decline is due to excessive data volume leading to a suboptimal data distribution relative to the validation set.

\paragraph{Augmented Training Data Enriches the Overall Data Distribution}
\begin{figure}[t]
  \centering
  \includegraphics[width=0.9\linewidth]{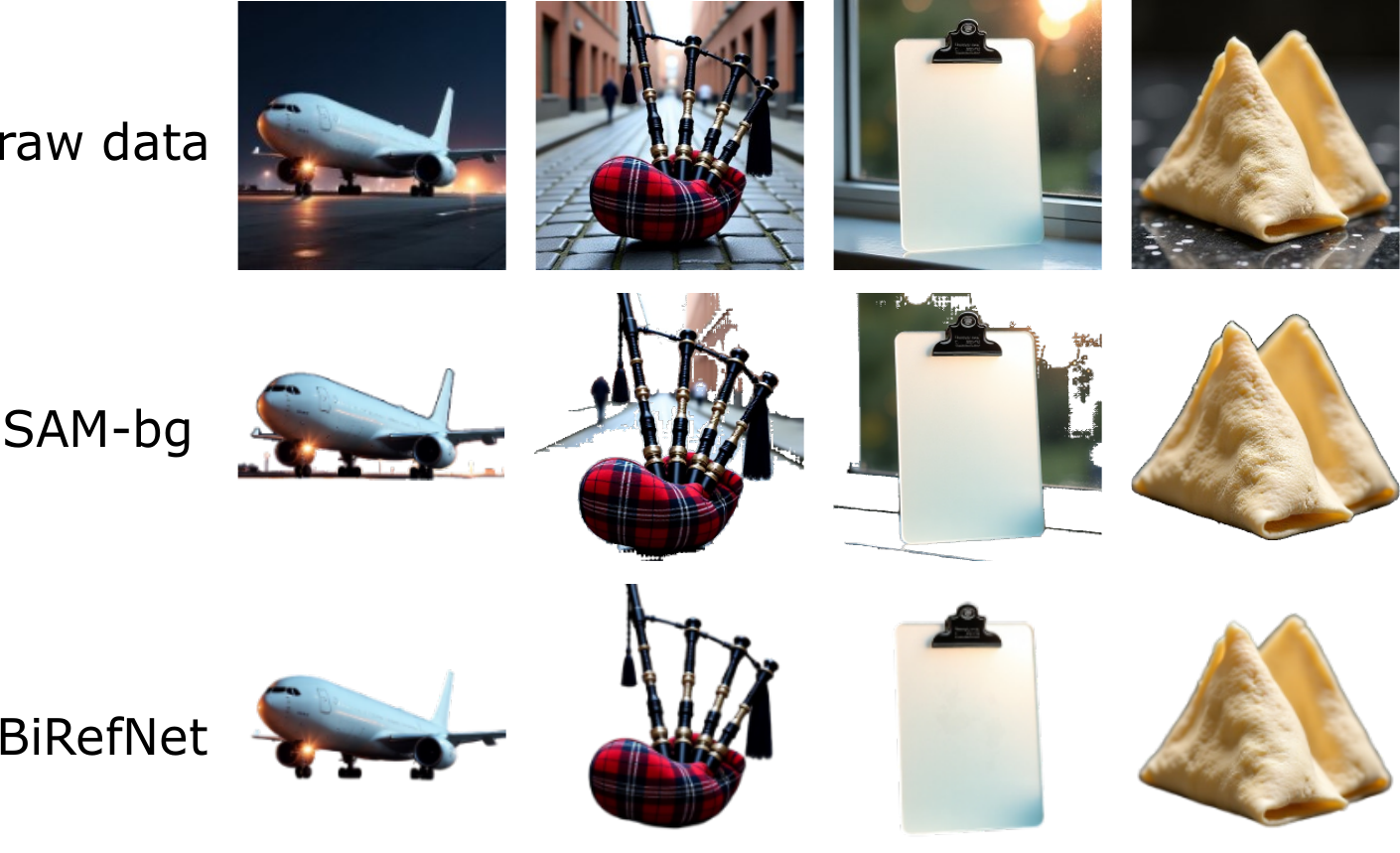}
  \caption{Examples of SAM-bg and BiRefNet. BiRefNet produces more precise segmentation masks than SAM-bg, demonstrating superior performance on generated images with both simple and complex backgrounds.}
  \label{fig:fig_segmentation}
\end{figure}
\begin{table}[t]
  \small
  \centering
  \setlength{\tabcolsep}{3mm} 
      \begin{tabular}{c|cccc}
      \toprule
       Method&$\text{AP}^{box}$&$\text{AP}^{mask}$&$\text{AP}_r^{box}$&$\text{AP}_r^{mask}$ \\
      \midrule
      none & 51.00     & 45.26    & 48.29    & 42.98 \\      
      Copy-Paste & 51.39  & 45.41  & 48.61  & 42.77 \\
      I-Agent  & \textbf{51.51}  & \textbf{45.51}  & \textbf{50.10} & \textbf{44.31} \\
    \bottomrule
    \end{tabular}
\vspace{-2mm}
\caption{Results of the integration of I-Agent. Integrating the I-Agent yields further improvements in model performance, indicating that it enhances the domain alignment of the data distribution.}
\vspace{-5mm}
\label{tab:IAgent}
\end{table}
\begin{table*}[t]
  \small
  \centering
  \setlength{\tabcolsep}{1mm}
    \begin{tabular}{r |c|cccccccc}
      \toprule
      Method & Backbone & AP\(^{box}\) & AP\(^{mask}\) & AP\(_r^{box}\) & AP\(_c^{box}\) & AP\(_f^{box}\) & AP\(_r^{mask}\) & AP\(_c^{mask}\) & AP\(_f^{mask}\) \\
      \midrule
      Copy-Paste~\cite{ghiasi2021simple} & EfficientNet-B7 & 41.6 & 38.1 & - & - & - & 32.1 & 37.1 & 41.9 \\
      Tan et al.~\cite{tan20201st} & ResNeSt-269  & - & 41.5 & - &- &- & 30.0 & 41.9 & 46.0 \\ 
      Detic~\cite{zhou2022detecting} & Swin-B & 46.9 &41.7 & 45.9 & -& -&  41.7 & - & - \\
      CenterNet2~\cite{zhou2021probabilistic} &  Swin-L & 47.5 & 42.3 & 41.4 & 47.5 & 50.1 & 36.8 & 43.1 & 43.8 \\
      \midrule
      X-Paste~\cite{zhao2023x} & Swin-L  & 50.9 & 45.4 & 48.7  & 50.7 & 52.2 & 43.8 & 45.8 & \textbf{45.5} \\
      DiverGen~\cite{fan2024divergen} & Swin-L  & 51.2 & 45.5 & \textbf{50.1} & \textbf{51.3} & 51.6 & \textbf{45.8} & 45.8 & 45.0 \\
      InstaGA  & Swin-L & \textbf{51.5} & \textbf{45.6} & \textbf{50.1} & \textbf{51.3} & \textbf{52.3} & 44.6 & \textbf{46.0} & \textbf{45.5} \\
    \bottomrule 
    \end{tabular}
\caption{Comparison on the LVIS val set, where bold indicates the best performance among Swin-L backbone methods.}
\label{tab:tab_main}
\end{table*}
\begin{table}[t]
  \small
  \centering
  \setlength{\tabcolsep}{1.3mm} 
  \begin{tabular}{c|cccc}
    \toprule
    {Segmentation method}&$\text{AP}^{box}$&$\text{AP}^{mask}$&$\text{AP}_r^{box}$&$\text{AP}_r^{mask}$ \\
    \midrule
    SAM-bg~\cite{fan2024divergen} & 50.62 & 44.78 & 47.49 & 42.13 \\
    BiRefNet &  \textbf{50.79} & \textbf{45.03} & \textbf{48.47} & \textbf{43.26}  \\
    \bottomrule
  \end{tabular}
  \caption{Comparison of using different segmentation methods. BiRefNet yields superior model performance, as it produces more accurate masks on generative images with complex backgrounds.}
  \label{tab:table_segmentation}
\end{table}

\begin{table}[t]
  \small
  \centering
  \setlength{\tabcolsep}{1pt} 
      \begin{tabular}{c|cccc}
      \toprule
      Metric & $\text{AP}^{box}$ & $\text{AP}^{mask}$ & $\text{AP}_r^{box}$ & $\text{AP}_r^{mask}$ \\
      \midrule
      CLIP score~\cite{zhao2023x} & 50.64 & 44.92 & 47.36 & 42.60 \\      
      CLIP inter-sim.~\cite{fan2024divergen} & 50.33 & 44.53 & 46.44 & 41.36 \\
      CLIP dual-sim. & \textbf{50.79} & \textbf{45.03} & \textbf{48.47} & \textbf{43.26} \\
      \bottomrule
    \end{tabular}
\vspace{-2mm}
\caption{Ablation of different filtration metrics. CLIP dual-similarity integrates the strengths of the two aforementioned methods to preserve high-quality images.}
\vspace{-5mm}
\label{tab:dual_clip}
\end{table}

      
\begin{table}[t]
  \small
  \centering
  \setlength{\tabcolsep}{2.5mm} 
      \begin{tabular}{c|cccc}
      \toprule
      Prompt Rethink&$\text{AP}^{box}$&$\text{AP}^{mask}$&$\text{AP}_r^{box}$&$\text{AP}_r^{mask}$ \\
      \midrule
      \xmark     & 50.41     & 44.59    & 45.64    & 40.48 \\      
      \cmark & \textbf{50.45}  & \textbf{44.65}  & \textbf{46.33}  & \textbf{41.09} \\
    \bottomrule
    \end{tabular}
\caption{Ablation of the function of Prompt Rethink. The Prompt Rethink mechanism boosts performance by refining prompts for low-quality generations.}
\label{tab:table_rethink}
\end{table}

\begin{table}[t]
  \small
  \centering
  \setlength{\tabcolsep}{3.3mm} 
      \begin{tabular}{c|cccc}
      \toprule
       Strategy&$\text{AP}^{box}$&$\text{AP}^{mask}$&$\text{AP}_r^{box}$&$\text{AP}_r^{mask}$ \\
      \midrule
       none     & 51.22     & 45.26    & 47.83    & 42.43 \\      
      SAM-box & \textbf{51.33}  & \textbf{45.43}  & \textbf{48.82}  & \textbf{43.59} \\
    \bottomrule
    \end{tabular}
\vspace{-2mm}
\caption{Ablation of the annotation strategies. The SAM-box strategy significantly boosts model performance by improving the accuracy of segmentation masks.}
\vspace{-5mm}
\label{tab:table_SAM-box}
\end{table}

Our analysis shows that while excessive generated data can increase diversity, it may also lead to a suboptimal data distribution. To address this, we propose that performance can be improved by augmenting the training set with generated data specifically designed to enrich the overall data distribution. To this end, our I-Agent generates a counterpart for each of the 99,388 training images. After filtering this pool to retain the top 20\% of samples, these new instances are combined with the original training data to create an enhanced source pool for Copy-Paste.

The results in Table~\ref{tab:IAgent} clearly support our argument. While standard Copy-Paste using training instances shows significant improvement, integrating this data with our I-Agent yields an additional gain of +0.12 box AP and +0.1 mask AP. This result demonstrates the effectiveness of our I-Agent in producing high-quality instances that contribute to a more robust data distribution. Our findings confirm that while diversity is important, actively enriching the data distribution is essential for achieving optimal performance.
\paragraph{Comparison with Previous Methods}
As summarized in Table~\ref{tab:tab_main}, our method yields substantial gains of +4.0 box AP and +3.3 mask AP over the CenterNet2 baseline~\cite{zhou2021probabilistic} and outperforms the leading model DiverGen~\cite{fan2024divergen} with +0.3 box AP and +0.1 mask AP. The outperformance is particularly compelling, demonstrating that the superior data diversity and data distribution successfully compensate for the quality trade-off introduced by LoRA to significantly speed up the generation process. This success stems from our dual-agent system, where the T-Agent generates highly diverse data and the I-Agent enriches the data distribution.

\subsection{Ablation Studies}
\paragraph{Ablation on Foreground Segmentation Methods}
We evaluate BiRefNet~\cite{zheng2024birefnet,bireftrt} and SAM-bg~\cite{fan2024divergen} on 480k images generated by Flux. As shown in Table~\ref{tab:table_segmentation}, BiRefNet achieves superior performance. The primary reason is that SAM-bg struggles with the non-blank backgrounds in our generated data. Qualitative results in Fig.~\ref{fig:fig_segmentation} further confirm the robust generalizability of BiRefNet, consistently producing accurate masks regardless of background complexity.

\paragraph{Ablation on Filtration Metrics}
To demonstrate the advantages of our proposed CLIP dual-similarity metric, we compare its performance against using either CLIP inter-similarity~\cite{fan2024divergen} or CLIP score~\cite{zhao2023x} alone as the filtration metric. As shown in Table~\ref{tab:dual_clip}, our metric achieves superior performance. It suggests that combining both semantic and visual quality checks allows for a more robust filtration of low-quality generated instances.

\paragraph{Effect of Prompt Rethink}
We assess the impact of our Prompt Rethink mechanism on a 480k generated image dataset. As shown in Table~\ref{tab:table_rethink}, incorporating this module yields an improvement in model performance. This gain is attributed to its core feedback loop, where the LLM iteratively refines prompts based on visual feedback from the diffusion model. This interactive process enhances both overall data utilization and the fidelity of the generated instances.
\paragraph{Effect of SAM-box Strategy} 
We compare the performance of training with original masks against that with masks refined by our SAM-box strategy. As shown in Table~\ref{tab:table_SAM-box}, SAM-box enhances performance. This improvement is important because original segmentation masks can become misaligned with instances in newly generated images. To address this, SAM-box uses bounding boxes as reliable prompts for SAM, allowing it to generate new and accurate masks, effectively correcting such boundary inaccuracies.

\section{Conclusion}
In this paper, we introduce InstaDA, a dual-agent framework. The T-Agent generates highly diverse data by pioneering a Prompt Rethink feedback loop between LLMs and diffusion models, while the I-Agent enriches the data distribution by augmenting the training set with domain-aligned generated data. Our framework incorporates precise segmentation via SAM-box, a novel CLIP dual-similarity quality metric, and an efficient ComfyUI-based implementation. InstaDA enhances data diversity while enriching the data distribution, thereby enabling effective large-scale generative data augmentation to boost model performance.

\bibliography{InstaDA}
\appendix
\twocolumn[
\begin{center}
\LARGE\bfseries InstaDA: Augmenting Instance Segmentation Data with Dual-Agent System\\
\vspace{1em}
Supplementary Material
\end{center}
\vspace{6em}
]

\section{Training Details}
For fair comparison and reproducibility, all our experiments are conducted on a server equipped with eight NVIDIA A800 and L20 GPUs, using CUDA 11.6 and PyTorch 1.13.1. Regarding the training time, the baseline model completes its training in approximately 1-2 days. With our proposed data augmentation, the training duration extends to 3-5 days, incurring a computational overhead that is comparable to that of DiverGen~\cite{fan2024divergen}. 
This increase is primarily due to the on-the-fly processing overhead of the Copy-Paste augmentation applied to each training batch.

\section{Efficiency Analysis}
\paragraph{Generative Configuration} The Text-Agent (T-Agent) employs two diffusion models: (1) Flux.1-dev~\cite{flux2024}, configured to generate images at a 512 $\times$ 512 resolution using the Euler sampler with a guidance scale of 9.5; and (2) Stable Diffusion 3.5-large~\cite{sd3.5}, which generates images at a 1024 $\times$ 1024 resolution, also using the Euler sampler with its default settings. The Image-Agent (I-Agent) utilizes Flux.1-dev with a ControlNet~\cite{zhang2023adding} strength of 0.7 for structural guidance. Crucially, the integration of LoRA~\cite{hu2022lora} allows us to reduce the sampling steps for all generative processes to just 8, significantly improving efficiency.

\paragraph{Efficiency Analysis and Comparison} Our generation process stands in stark contrast to the prior work DiverGen~\cite{fan2024divergen}, which relies on a computationally intensive pipeline involving two separate stages. The process of DiverGen involves a 100-step base generation followed by a 50-step upsampling and refinement stage, totaling 150 sampling steps. While this approach may enhance image quality, it incurs a prohibitive temporal cost, especially for large-scale data generation. In contrast, our methodology, accelerated by LoRA, accepts a marginal reduction in image quality in exchange for a dramatic acceleration. By reducing the process to a single stage of only 8 sampling steps, we eliminate this bottleneck, thereby making large-scale generative augmentation feasible. As evidenced in Table~\ref{tab:time_compare}, this approach results in a substantial speedup, accelerating the data generation process by a factor of \textbf{6.8} to \textbf{25} times compared to DiverGen.
\begin{table}[t]
  \small
  \centering
  \setlength{\tabcolsep}{5.5mm} 
      \begin{tabular}{c|cc}
      \toprule
       Method & Runtime (s)& Total Steps \\
      \midrule
       DiverGen  & 60     &  150 \\  
       T-Agent (Flux) & \textbf{2.4} &  \textbf{8} \\
       T-Agent (SD3.5) & 4 & \textbf{8} \\
       I-Agent (Flux) & 8.7 & \textbf{8} \\
    \bottomrule
    \end{tabular}
\caption{Comparison of average generation time and total steps per image for InstaDA and DiverGen. InstaDA significantly accelerates the image generation process, thereby enhancing its practical applicability.}
\label{tab:time_compare}
\end{table}

\begin{table}[t]
  \small
  \centering
  \setlength{\tabcolsep}{1.3mm} 
      \begin{tabular}{c|cccc}
      \toprule
       Strategy&$\text{AP}^{box}$&$\text{AP}^{mask}$&$\text{AP}_r^{box}$&$\text{AP}_r^{mask}$ \\
      \midrule
      Text similarity threshold & 50.87  & 45.01  & 46.61  & 41.61 \\
      Proportional filtration & \textbf{51.51}  & \textbf{45.51}  & \textbf{50.10}  & \textbf{44.31} \\
    \bottomrule
    \end{tabular}
\caption{Ablation study on I-Agent filtration strategies. Proportional filtration (with a 30\% ratio) significantly outperforms the baseline using a text similarity threshold of 21.}
\label{tab:filtration_ablation}
\end{table}

\begin{table*}[t]
  \small
  \centering
  \setlength{\tabcolsep}{4.2mm}
      \begin{tabular}{c|c|c|cccc}
      \toprule
      \ Gen Data& Text Similarity Threshold & Image Similarity Threshold&$\text{AP}^{box}$&$\text{AP}^{mask}$&$\text{AP}_r^{box}$&$\text{AP}_r^{mask}$ \\
      \midrule
      480k & 0.25 & 0.6 & 50.41  &  44.59 & 45.64 & 40.48 \\
      480k & 0.23 & 0.6 & 50.52  &  44.80 & 47.41 & 42.41 \\
      480k & 0.22 & 0.6 & 50.73  & 44.73 & 48.33 & 42.51 \\
      480k & 0.21 & 0.6 & \textbf{50.79}  &  \textbf{45.03} & \textbf{48.47} & \textbf{43.26} \\
      480k & 0.20 & 0.6 & 50.49  &  44.86 & 47.17 & 42.24 \\
      480k & 0 & 0.6 & 50.33  &  44.53 & 46.44 & 41.36 \\
      480k & 0.21 & 0 & 50.64  &  44.92 & 47.36 & 42.60 \\
    \bottomrule
    \end{tabular}
\caption{Ablation study on the thresholds for the CLIP dual-similarity metric. The results demonstrate that a text similarity threshold of 0.21 and an image similarity threshold of 0.6 achieve the best performance across all evaluation metrics.}
\label{tab:compare_clip_score}
\end{table*}
\section{Ablation Studies}
\paragraph{Effect of the I-Agent Proportional Filtration}
We compare our proportional filtration against a method based on a text similarity threshold. As shown in Table~\ref{tab:filtration_ablation}, the threshold-based approach degrades performance. In contrast, our proportional filtration method yields a notable performance improvement. This degradation occurs because threshold-based, per-instance filtering is incompatible with per-image Copy-Paste~\cite{ghiasi2021simple}, which can lead to pasting incomplete instance sets from a source image. Conversely, our proportional filtration operates at the image level, aligning the filtration unit with the Copy-Paste unit. This alignment preserves the completeness of instance groups, which directly translates to the performance improvement.
\paragraph{Ablation on Thresholds for CLIP dual-similarity}
Table~\ref{tab:compare_clip_score} presents our ablation study to determine the optimal parameters for our CLIP dual-similarity metric: a text similarity threshold of 0.21 and an image similarity threshold of 0.6. This combination yields the best performance because it strikes an optimal balance between preserving data diversity and maintaining high domain fidelity.
\begin{table}[t]
  \small
  \centering
  \setlength{\tabcolsep}{3.3mm} 
      \begin{tabular}{c|cccc}
      \toprule
       Proportion &$\text{AP}^{box}$&$\text{AP}^{mask}$&$\text{AP}_r^{box}$&$\text{AP}_r^{mask}$ \\
      \midrule
       30     & 51.18     & 45.37    & 47.73    & 43.07 \\      
      20 & \textbf{51.33}  & \textbf{45.43}  & \textbf{48.82}  & \textbf{43.59} \\
      10 & 50.07  & 45.21  & 48.03  & 42.56 \\
    \bottomrule
    \end{tabular}
\caption{Ablation on the proportional filtration ratio $k$. The results show that a 20\% data expansion ($k=20$) achieves the best performance across all evaluation metrics.}
\label{tab:k_proportion}
\end{table}

\paragraph{Ablation on Proportional Filtration Ratio} We conduct an ablation study to find the optimal proportional filtration ratio, denoted by the parameter $k$. This parameter specifies the data expansion percentage, where a value of $k=20$ corresponds to a 20\% expansion. As illustrated in Table~\ref{tab:k_proportion}, the performance peaks at $k=20$. This suggests that a 20\% data expansion achieves the most effective overall data distribution by introducing adequate diversity. Accordingly, we set $k=20$ for our method in all experiments.



\end{document}